\documentclass[runningheads]{llncs}
\usepackage{graphicx}
\usepackage{amsmath}

\usepackage{algorithm}
\usepackage{algorithmicx}
\usepackage{algpseudocode}
\usepackage{subfigure}
\usepackage{multirow}

\usepackage{CJKutf8}
\usepackage{makecell}
\usepackage{booktabs}
\usepackage[misc]{ifsym}
\usepackage[bookmarks=false,colorlinks,linkcolor=blue,anchorcolor=blue,citecolor=blue]{hyperref} 

\newcommand{\etal}{\textit{et al}.~}

\begin{document}
\title{Finding Better Subword Segmentation for Neural Machine Translation}
\titlerunning{Finding Better Subword Segmentation for Neural Machine Translation}
%
\author{Yingting Wu\inst{1,2} \and Hai Zhao\inst{1,2}$^\text{(\Letter)}$}
\authorrunning{Y. Wu and H. Zhao}
%
\institute{Department of Computer Science and Engineering, Shanghai Jiao Tong University \and
Key Laboratory of Shanghai Education Commission for Intelligent Interaction
\\and Cognitive Engineering, Shanghai Jiao Tong University, Shanghai, 200240, China
\email{wuyingting@sjtu.edu.cn, zhaohai@cs.sjtu.edu.cn}}

\maketitle              
\begin{abstract}
For different language pairs, word-level neural machine translation (NMT) models with a fixed-size vocabulary suffer from the same problem of representing out-of-vocabulary (OOV) words. The common practice usually replaces all these rare or unknown words with a $\langle$UNK$\rangle$ token, which limits the translation performance to some extent. Most of recent work handled such a problem by splitting words into characters or other specially extracted subword units to enable open-vocabulary translation. Byte pair encoding (BPE) is one of the successful attempts that has been shown extremely competitive by providing effective subword segmentation for NMT systems. In this paper, we extend the BPE style segmentation to a general unsupervised framework with three statistical measures: frequency (FRQ), accessor variety (AV) and description length gain (DLG). We test our approach on two translation tasks: German to English and Chinese to English. The experimental results show that AV and DLG enhanced systems outperform the FRQ baseline in the frequency weighted schemes at different significant levels.

\end{abstract}

\section{Introduction}
Neural Machine Translation \cite{Bahdanau15,kalchbrenner-blunsom,Sutskever14} has achieved remarkable performance in recent years. The NMT system is commonly based on a word-level model with a restricted vocabulary of most frequent words. All the infrequent or unseen words are simply replaced with a special $\langle$UNK$\rangle$ token at the cost of decreasing translation accuracy. In order to better handle the problems of OOV words, recent work proposed the character-based \cite{chung-cho-bengio2016,costajussa-fonollosa:2016:P16-2,lee2017fully} or word-character hybrid models \cite{luong-manning2016}. Considering that character is not a sufficiently good minimal unit to compose word representation, it is of great value to find a more meaningful representation unit between word and character, i.e., subword.

The benefits of using variable-length subword representation in NMT are two-fold. First, it can effectively decrease the vocabulary size of the whole training set by encoding rare words with sequences of subword units. Second, it is also possible to translate or generate unseen words at inference. Currently, BPE \cite{gage1994new} has been applied commonly in NMT called subword NMT \cite{sennrich-haddow-birch} that enables open-vocabulary translation. It is a frequency-based data compression algorithm that most of the high-frequency words are maintained and the infrequent ones are segmented into subwords. As the widely used BPE segmentation only uses frequency as its measure and does not rely on any morphological or linguistic knowledge, it can be regarded as a computationally motivated method. Some other work segmented words based on morphological information to obtain the representation of OOV words \cite{ataman2017linguistically,botha2014compositional}, which can be called linguistically motivated method. However, the availability of morphological resources for different languages cannot be always guaranteed. Thus, this work will focus on the general computationally motivated subword segmentation for NMT. 

The development of Chinese word segmentation has played an important role in the natural language processing tasks \cite{neuralseg,fastseg,empseg}. Zhao and Kit \cite{zhao2008empirical} proposed a general unsupervised word segmentation framework for Chinese which consists of two collocative components, decoding algorithm and substring goodness measure. Their work shows that the substring frequency (FRQ) is not the best goodness measure for Chinese word segmentation. Since Chinese is an ideographic language without any explicit word boundaries, we can similarly adapt the reported better measures to the subword segmentation for alphabetic languages such as English by considering their words are written in character sequences without explicit character boundaries. Thus, we regard the BPE style segmentation as a general decoding algorithm, and enable it to work with another two advanced goodness measures, accessor variety (AV) \cite{feng2004accessor} and description length gain (DLG) \cite{kitt1999unsupervised} in the hope of further enhancing the current NMT. We compare the effects of different measures with experiments on the German-English and Chinese-English translation tasks. The evaluations on the test sets prove that the extended BPE-style segmentation methods with the frequency weighted AV and DLG improve the FRQ baseline at different significant levels.

\section{Neural Machine Translation}
In this paper, we closely follow the neural machine translation model proposed by  Bahdanau \etal \cite{Bahdanau15}, which is mainly based on a neural encoder-decoder network with attention mechanism \cite{Bahdanau15,luong-pham-manning}.

The encoder is a bidirectional recurrent neural network (RNN) with gated recurrent unit (GRU) or long short-term memory (LSTM) unit. The forward RNN reads a source sentence $x=(x_1, ..., x_m)$ from left to right and calculates a forward sequence of hidden states $(\overrightarrow{h}_1, ..., \overrightarrow{h}_m)$. Similarly, the backward RNN reads the source sentence inversely and learns a backward sequence of hidden states $(\overleftarrow{h}_1, ..., \overleftarrow{h}_m)$. The hidden states of both directions are concatenated to obtain the annotation vector $h_i=[\overrightarrow{h}_i, \overleftarrow{h}_i]^T$ for each word in the source sentence. 

The decoder is a forward RNN initialized with the final state of the encoder. In the decoding phase, a target sentence $y = (y_1,...,y_n)$ is generated step by step. The conditional translation probability can be formulated as follows.
\begin{equation*}
	p(y_j|y_{<j}, x) = q(y_{j-1}, s_j, c_j), 
\end{equation*}
where $s_j$ and $c_j$ denote the decoding state and the source context at the $j$-th time step respectively. Here, $q(\cdot)$ is the softmax layer and $y_{<j}=(y_1, ..., y_{j-1})$.
The context vector $c_j$ is calculated as a weighted sum of the source annotations according to attention mechanism, where $c_j = \sum_{i=1}^{m}\alpha_{ji}h_i$.
The alignment model $\alpha_{ji}$ defines the probability that how well $y_j$ is aligned to $x_i$, which can be a single layer feed-forward neural network.

\section{Unsupervised Subword Segmentation}

Extracting substrings inside a word can be regarded as a segmentation process over character sequences, which is similar to splitting words over Chinese character sequences (e.g., Chinese sentences). Therefore, we borrow the idea from Zhao and Kit \cite{zhao2008empirical} who proposed a generalized framework for unsupervised Chinese word segmentation including two collocative modules, a decoding algorithm and an alternative goodness measure.  They use a top-down decoding method that starts from the sentence level and then searches for the best segmentation for a particular text according to the given goodness function. Their empirical assessment on Chinese used the frequency of substrings as baseline among all goodness measures, which indicates that frequency is not an optimal segmentation criterion for Chinese. Motivated by this, we propose a combined framework of the extended BPE-style segmentation and some advanced substring measures for better subword representation in NMT.

\subsection{Generalized BPE Segmentation}

The BPE used in subword NMT is a bottom-up method, where the initial state of each word is a sequence of single characters and consecutive substring pairs with highest frequency are merged iteratively to compose subwords. In this section, we introduce a generalized BPE segmentation framework with arbitrary goodness measures besides frequency. The details are described in Algorithm \ref{algor1}.\footnote{The source code has been released at \url{https://github.com/Lindsay125/gbpe}.}

 Given the corresponding segmentation measure, if a substring pair (`a', `b') has the highest goodness score among all the candidate pairs, every occurrence of this pair will be replaced by a new symbol `ab'. The merge operations are performed on the whole training corpus and the vocabulary size can be controlled by the number of merge operations $N$. The final vocabulary size is approximately equal to merge times plus the number of character types in the corpus. According to the merge list of substring pairs learned from training data, we can further apply the merge operation to the development set and test set. Each subword except the end of word is attached to a special token such as ``@@'' for the sake of restoring the segmented words after translation. 

\begin{algorithm}[H]
	\caption{Generalized BPE segmentation}
	\label{algor1}
	\begin{algorithmic}[1]
		\Require the training corpus $D$, merge times $N$, goodness measure $g$
		\Ensure the segmented text $D'$, merge list $V$
		\State The training corpus $D$ is initialized as a set of character sequences in which every word is split into a sequence of characters, and the merge list $V$ is set empty.
		\State Given the current segmentation state on $D$, calculate the goodness scores of all the distinct successive substring pairs according to the goodness measure $g$; 
		\State Search for the highest scored consecutive pair, add it to $V$ and merge all the occurrences of such pairs on $D$;
		\State If the merge times reaches $N$, the algorithm ends and returns $D'$ and $V$.
		\State Otherwise, go to Step 2.
	\end{algorithmic}	
\end{algorithm}

\subsection{Goodness Measures}
Given a word $W$ with $n$ characters, its segmentation $S$ can be denoted as a subword sequence $s_1s_2\cdots s_m$ ($m\leq n$). For each merge iteration, every subword $s_i$ (the concatenation of a consecutive pair) is assigned a goodness score $g(s_i)$ for how likely it is an independently translatable item within the whole word. In this study, we examine three types of goodness measures.

\subsubsection{Frequency of Substring }
The \textit{frequency of substring} (FRQ) serves as the baseline in the Chinese word segmentation system of Zhao and Kit \cite{zhao2008empirical}. Its basic idea is to compare the frequency of two partially overlapped character $n$-grams and then the shorter substring with lower or equal frequency is discarded as a redundant candidate. We define the corresponding goodness score as $g_{FRQ}(s_i)$, which is the count of word types in the vocabulary that contain $s_i$. For efficiency, only those substrings that occur over once are considered in the candidate list.

\subsubsection{Accessor Variety}
The \textit{accessor variety} (AV) proposed by Feng \etal \cite{feng2004accessor} is a measure to evaluate how likely a substring can be a relatively independent word, which is reported to be good at addressing low frequency words. Given a particular substring, the main idea of AV is that if the type of successive tokens with respect to the corresponding substring increases, it is more likely to be at boundary.  Formally, AV is defined as the minimum of $L_{av}(s_i)$ and $R_{av}(s_i)$.
\begin{equation*}
	g_{AV}(s_i) = AV(s_i) = \min{\{L_{av}(s_i), R_{av}(s_i)\}},
\end{equation*}
where the left AV $L_{av}(s_i)$ is the number of distinct predecessor tokens of $s_i$ and the right AV $R_{av}(s_i)$ records the number of $s_i$'s different successor tokens.

\subsubsection{Description Length Gain}
The \textit{description length} of a corpus $X$ is defined as the Shannon-Fano code length for the corpus \cite{kit1998goodness}. It can be calculated as
\begin{equation*}
		DL(X) = - |X|\sum_{x \in V_X} \hat p(x) \log \hat p(x)  = -\sum_{x \in V_X} c(x) \log \frac {c(x)}{ |X|},
\end{equation*}
where $V_X$ is the token vocabulary of $X$, $c(x)$ is the count of token $x$ in $X$, $\hat p(x)=c(x)/|X|$ and $|X|$ represents the total token count in $X$.

Kit and Wilks \cite{kitt1999unsupervised} proposed to use the \textit{description length gain} (DLG) for lexical acquisition on word boundary prediction tasks and showed the effectiveness of this goodness measure. The DLG of a particular substring $s_i$ in $X$ is then defined as the description length change while we substitute $s_i$ with an index $r$ and take a note of this operation at the end of $X$, i.e.,
\begin{equation*}
	g_{DLG}(s_i) = DLG(s_i) = DL(X)-DL(X[r\rightarrow s_i]\oplus s_i), 
\end{equation*}
where $X[r\rightarrow s_i]$ is the resultant corpus by replacing all occurrences of $s_i$ with $r$ throughout $X$ and $\oplus$ represents the concatenation of two strings with a delimiter.

\subsubsection{Frequency Weighted Schemes}\label{sec3}

Note that parallel corpus for NMT typically holds a large vocabulary with noise, all of these statistical goodness measures may be biased by the data noise. Especially for FRQ and AV, which are only based on type statistics,  their calculation will be misled by too many word types even though each word type has low frequency. Considering that high frequency words usually correspond to those reliable ones with regular forms, we also introduce three frequency weighted variants,\footnote{Though DLG is already frequency weighted as its definition,  the proposed extra frequency weight is empirically verified effective from our preliminary experiments.}
\begin{equation*}
	g'_{FRQ}(s_i)=g_{FRQ}(s_i)\sum_{\forall w, s_i\in w}f(w), 
\end{equation*}
\begin{equation*}	
	g'_{AV}(s_i)=g_{AV}(s_i)\sum_{\forall w, s_i\in w}f(w),~~~~ 
	g'_{DLG}(s_i)=g_{DLG}(s_i)\sum_{\forall w, s_i\in w}f(w), 
\end{equation*}
where $f(w)$ is the frequency of word $w$ in the corpus. Note that here our FRQ'-BPE will slightly differ from the BPE in \cite{sennrich-haddow-birch} whose goodness score is directly computed through counting $s_i$ in the entire corpus.

\section{Experiments}
\subsection{Setup}
Our experiments will be performed on two typical language pairs, German to English and Chinese to English. The translation quality is evaluated by the 4-gram case-sensitive BLEU \cite{papineni-EtAl:2002:ACL} and we use sign test \cite{collins2005clause} to test the statistical significance of our results.

For German-English task, the evaluations are based on data from TED talks corpora of the IWSLT 2014 evaluation campaign \cite{IWSLT14}. We tokenize all the training data with the script of Moses\footnote{\url{https://github.com/moses-smt/mosesdecoder/blob/master/scripts/tokenizer}} and remove sentences longer than 50 words. The training set comprises of about 153$K$ sentence pairs with 100$K$ German words and 50$K$ English words. The development set consists of 6969 sentence pairs which is randomly extracted from training data. The test set is a concatenation of dev2010, dev2012, tst2010, tst2011 and tst2012 which results in 6750 sentence pairs. As German and English share the similar alphabet and have a large overlap of vocabulary, the learning of subwords is performed on the union of the source and target corpora as suggested by  Sennrich \etal \cite{sennrich-haddow-birch}. 

For Chinese-English task, we use the News Commentary v12 dataset from the news translation shared task of WMT 2017 \cite{bojar-EtAl:2017:WMT1} which consists of 227$K$ parallel sentence pairs. The development set and test set are randomly selected from the whole corpora with 3000 sentence pairs respectively, and the remaining part is served as the training set. Since Chinese and English have totally different character vocabularies, the subword (word for Chinese) learning is performed on the source and target corpora separately and we also try different learning strategies of segmentation on both sides. 

During training, we use LSTM units for both encoder and decoder. Each direction of the LSTM encoder and decoder are with 256 dimensions. The word embedding and the attention size are both set to 256.  The model is trained using Adam optimizer with the initial learning rate of 0.001. The batch size is set to 32 for German-English task and 64 for Chinese-English task. We train each model for 40 epochs and halve the learning rate every 10 epochs. The training set is reshuffled at the beginning of each epoch. During decoding, greedy search and beam search with size 10 are both performed to optimize the performance. 

\begin{table}[!htbp]
	\centering
	\caption{BLEU scores with normal AV and DLG on German-English test set}
	\begin{tabular}{l|c|c|c}
		\hline
		\multicolumn{1}{c|}{\multirow{2}{*}{\textbf{Method}}} & \multicolumn{3}{c}{\textbf{Merge Times}} \\
		\cline{2-4}         & \multicolumn{1}{c|}{10$K$} & \multicolumn{1}{c|}{20$K$} & \multicolumn{1}{c}{30$K$} \\
		\hline
		\textbf{FRQ-BPE} & 27.55 & 27.94 & \textbf{28.29} \\
		\textbf{AV-BPE}  & 27.42 & 27.96 & 27.60 \\
		\textbf{DLG-BPE} & 20.95 & 20.83 & 21.03 \\
		\hline		
	\end{tabular}
	\label{tab1}
\end{table}

\subsection{Results}

Table \ref{tab1} reports the translation performance after applying FRQ-BPE, DLG-BPE and AV-BPE segmentation on the German-English task with a beam size of 10. From Table \ref{tab1}, we notice that FRQ-BPE and AV-BPE outperform DLG-BPE at different merge times (vocabulary size) with a large margin and FRQ-BPE behaves the best among the three measures. Thus, we consider combining the corpus-level word frequency with the three goodness measures to further enhance the performance as we disscussed in Section \ref{sec3}.

\begin{table}[htbp]
	\centering
	\caption{BLEU scores with frequency-weighted schemes on German-English test set}
	\begin{tabular}{l|c|c|c|l|l|c}
		\hline
		\multicolumn{1}{c|}{\multirow{2}[4]{*}{\textit{\textbf{N}}}} & \multicolumn{2}{c|}{\textbf{FRQ$'$-BPE}} & \multicolumn{2}{c|}{\textbf{AV$'$-BPE}} & \multicolumn{2}{c}{\textbf{DLG$'$-BPE}} \\
		\cline{2-7}          & \multicolumn{1}{l|}{\textbf{Greedy}} & \multicolumn{1}{l|}{\textbf{Beam}} & \multicolumn{1}{l|}{\textbf{Greedy}} & \multicolumn{1}{l|}{\textbf{Beam}} & \multicolumn{1}{l|}{\textbf{Greedy}} & \multicolumn{1}{l}{\textbf{Beam}} \\
		\hline
		10$K$   & 27.36 & 28.86 & 27.72 & 29.02 & 27.56 & 29.25 \\
		15$K$  & 27.42 & 28.92 & 27.56 & 29.25 & 27.75 & 29.30 \\
		20$K$   & 27.51 & 28.99 & 27.65 & \textbf{29.46}$^{++}$ & \textbf{27.76}$^{+}$ & 29.36 \\
		25$K$   & 27.22 & 28.81 & 27.69 & 29.32 & 27.72 & 29.34 \\
		30$K$   & 27.06 & 28.72 & 27.52 & 29.09 & 27.46 & 29.03 \\
		\hline
	\end{tabular}%
	\label{tab2}
\end{table}%

\begin{figure}[!htbp] 
	\centering 
	\includegraphics[width=0.5\linewidth]{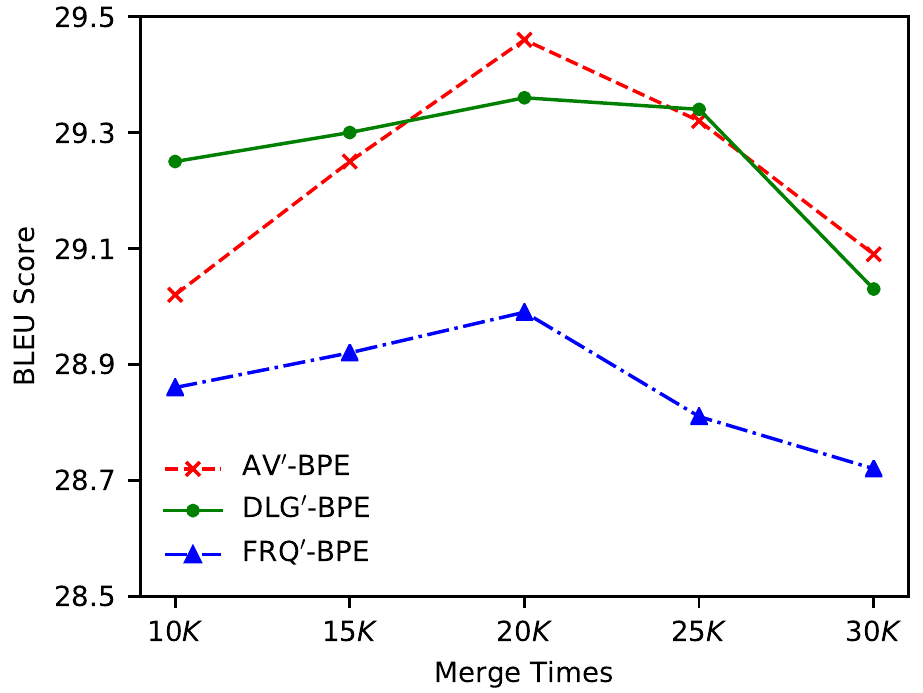} 
	\caption{The curves of BLEU scores with different merge times and  segmentation criteria on German-English test set (beam size=10).}\label{fig1} 
\end{figure}

Table \ref{tab2} reports BLEU scores with different frequency weighted goodness measures and corresponding merge times on the German-English test set.\footnote{``++'' indicates that the corresponding BLEU is significantly better than the best score of FRQ$'$-BPE at the significant level p $<$ 0.01, ``+'': p $<$ 0.05.} For greedy search, both DLG$'$-BPE and AV$'$-BPE achieve better performance than FRQ$'$-BPE and the best score is obtained with DLG$'$-BPE of 20$K$ merge operations. For beam search, AV$'$-BPE and DLG$'$-BPE outperform FRQ-BPE$'$ with improvement of 0.47 and 0.37 BLEU points when $N=20K$ respectively. Figure \ref{fig1} plots the variation curves of BLEU scores with different goodness measures against the merge times $N$ when the beam size is 10. From Figure \ref{fig1}, we observe that the peak value of BLEU with different criteria appears at the 20$K$ merge times. In general, AV$'$-BPE and DLG$'$-BPE are comparable with each other and both superior to FRQ$'$-BPE with a margin. 

\begin{figure}
	\centering 
	\subfigure[Average number of segmented words in each sentence.]{
		\includegraphics[width=0.45\columnwidth]{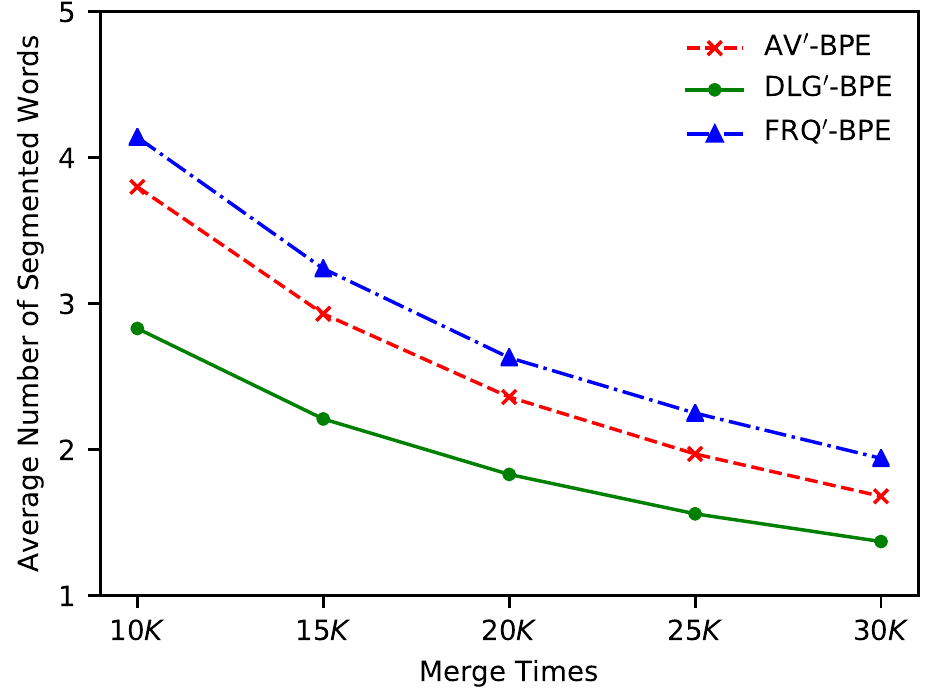} 
		\label{fig2:a}	
	}
	\subfigure[Average sentence length increase.]{
		\includegraphics[width=0.45\columnwidth]{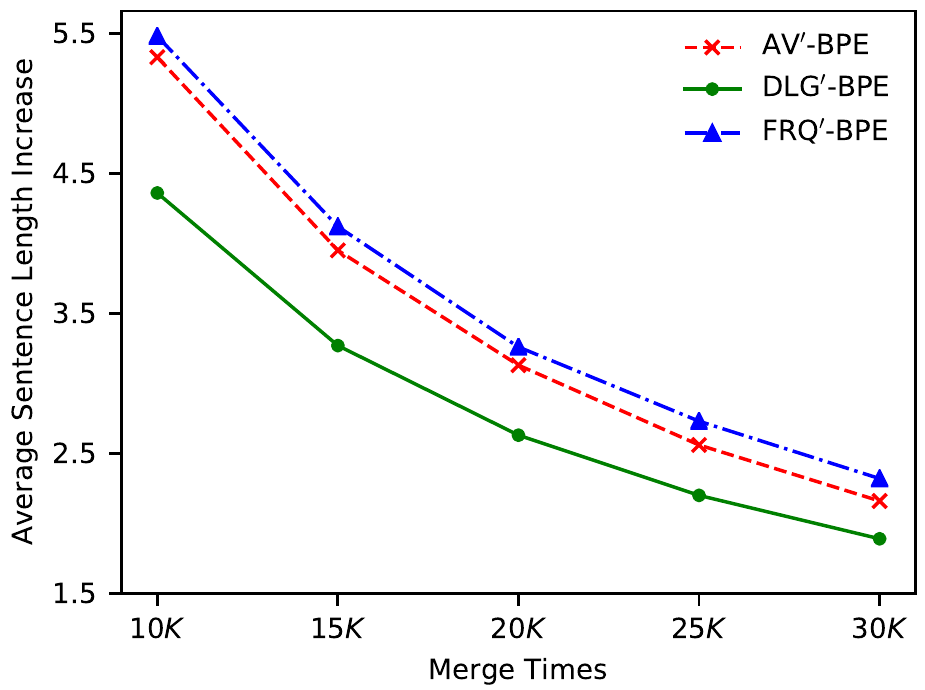} 
		\label{fig2:b}
	}
	\caption{Statistics of words and sentence after segmentation.}
	\label{fig2} 
\end{figure}

We also calculate and plot the average number of segmented words in each sentence and average sentence length increase after subword segmentation with different criteria on German-English task in Figure \ref{fig2}. Figure \ref{fig2:a} shows that, with the same merge times, AV$'$-BPE and DLG$'$-BPE maintain more original word forms than FRQ$'$-BPE within each sentence. The sentence length is an important factor during training as too long sequences will slow down the training and NMT shows a weak performance on long sentences. Figure \ref{fig2:b} also shows that both DLG$'$-BPE and AV$'$-BPE can better control the sequence length than FRQ$'$-BPE. For both Figure \ref{fig2:a} and \ref{fig2:b}, the curve of AV$'$-BPE with best performance is between that of DLG$'$-BPE and FRQ$'$-BPE, which means that it is crucial to find the proper granularity for segmentation. 

\begin{table}[htbp]
	\centering
	\caption{BLEU scores on Chinese-English test set (beam size=10)}
	\begin{tabular}{l|c|c|c}
		\hline
		\multicolumn{1}{c|}{\multirow{2}{*}{\textbf{Source}}} & \multicolumn{3}{c}{\textbf{Target}} \\
		\cline{2-4}         & \multicolumn{1}{l|}{\textbf{FRQ$'$-BPE}} & \multicolumn{1}{l|}{\textbf{AV$'$-BPE}} & \multicolumn{1}{l}{\textbf{DLG$'$-BPE}} \\
		\hline
		\textbf{FRQ$'$-BPE} & 20.07 &20.27 & 20.31 \\
		\textbf{AV$'$-BPE}  & 19.96 & 20.23 & 20.13 \\
		\textbf{DLG$'$-BPE} & 20.70 & 20.45 & \textbf{\quad20.72}$^{++}$ \\
		\hline		
	\end{tabular}
	\label{tab4}
\end{table}

\begin{figure}[!htbp] 
	\centering 
	\includegraphics[width=0.5\linewidth]{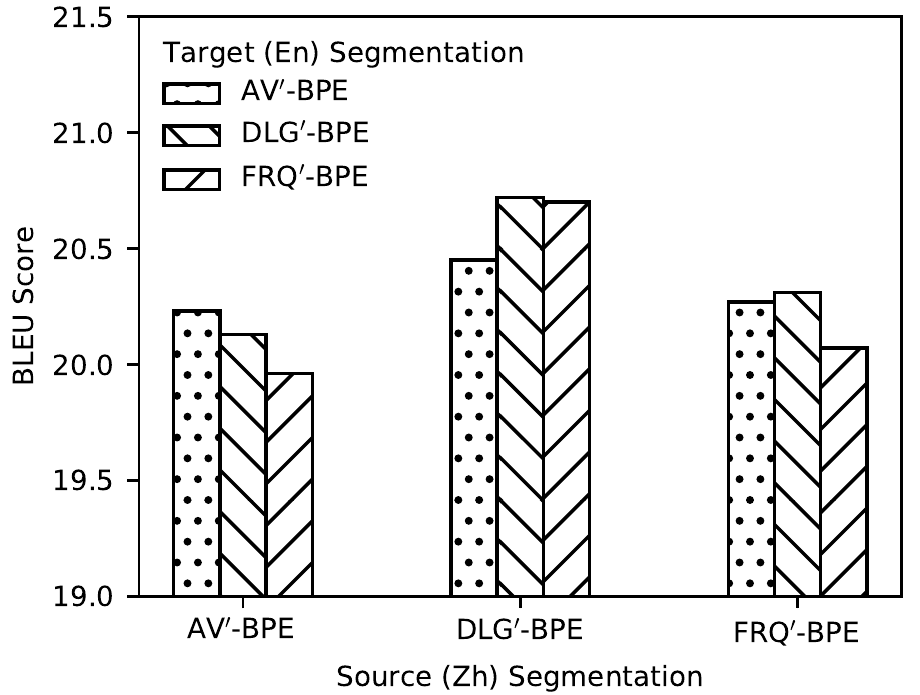} 
	\caption{The bar chart of BLEU scores with different segmentation criteria on Chinese-English test set (beam size=10).}\label{fig3} 
\end{figure}

In Table \ref{tab4}, we report the BLEU scores with different segmentation measures on the test set of Chinese-English task. Here, the subword segmentation learning is performed separately on the source and target sides with 30$K$ merge operations which is tuned the same way as German-English task. We also try different goodness measures on each side at the same time. It can be found that DLG$'$-BPE on the both sides significantly outperforms FRQ$'$-BPE with 0.65 BLEU points. Figure \ref{fig3} shows the translation performance comparison of different goodness measures when the source segmentation is fixed. It can be seen that DLG$'$-BPE does much better at the source segmentation for Chinese than the other two, while AV$'$-BPE and FRQ$'$-BPE are on a par with each other.

\begin{table}[h]
	\centering
	\small
	\caption{German-English and Chinese-English translation examples.}
	\begin{tabular}{l|l}
		\Xhline{1.2pt}
		Source &   ...und wozu wir als erstaunlich wissbegierige Spezies fähig sind .\\
		\hline
		Reference  &  ...and all that we can be as an astonishingly inquisitive species .\\
		\hline
		AV$'$-BPE  &...und wo$|$zu wir als erstaunlich wiss$|$beg$|$ier$|$ige spezies fähig sind . \\
		&  ...and what we are , as an aston$|$ish$|$ingly inven$|$tive species .\\
		\hline 
		DLG$'$-BPE  & ...und wo$|$zu wir als erstaunlich wis$|$sbe$|$gi$|$erige spezies fähig sind . \\
		&  ...and what we do as an ast$|$ound$|$ing species .\\
		\hline
		FRQ$'$-BPE  & ...und wo$|$zu wir als erstaun$|$lich wiss$|$beg$|$ier$|$ige spezies fähig sind .\\
		&   ...and what we are as an amazing$|$ly in$|$walk$|$able species .    \\

		\Xhline{1.2pt}
		Source &\begin{CJK}{UTF8}{gbsn}我们应当在多大程度上担心社会不平等呢~？\end{CJK} \\
		\hline
		Reference & How much should we worry about inequality ?\\
		\hline
		AV$'$-BPE & \begin{CJK}{UTF8}{gbsn}我们$|$应当$|$在多大程度上$|$担心$|$社会$|$不平等$|$呢~？\end{CJK}\\
		& What kind of social inequality we should be ?\\
		\hline
		DLG$'$-BPE & \begin{CJK}{UTF8}{gbsn}我们$|$应当$|$在$|$多大程度上$|$担心$|$社会$|$不平等$|$呢 ~？\end{CJK}\\
		& How much should we worry about social inequality ?\\
		\hline
		FRQ$'$-BPE &\begin{CJK}{UTF8}{gbsn}我们$|$应当$|$在$|$多$|$大程度$|$上$|$担心$|$社会$|$不$|$平等$|$呢~？\end{CJK}\\
		& How else should we worry about social inequality ?\\ 
		\hline
		
	\end{tabular}
	\label{tab5}
\end{table}

\subsection{Translation Examples}

Table \ref{tab5} shows some translation examples with different segmentation criteria on the test sets of German-English and Chinese-English tasks.  We use the ``$|$'' token to mark the splitting points within the words (sentences for Chinese).

In the first case, AV$'$-BPE performs best by translating ``\emph{erstaunlich wissbegierige}'' to an acceptable alternative ``\emph{astonishingly inventive}'' compared to the ground truth ``\emph{astonishingly inquisitive}''. DLG$'$-BPE omits the translation of ``\emph{wissbegierige}'' probably as the segmentation of this word is more confusing than that of the other two methods. FRQ$'$-BPE behaves not badly in the interpretation of ``\emph{erstaunlich}'' but generates a non-existing word ``\emph{inwalkable}''. 

\begin{CJK}{UTF8}{gbsn} In the second case, DLG$'$-BPE shows the better translation quality than the other two. The critical difference in splitting phrase ``在多大程度上'' (\emph{by what a degree}, \emph{how much})  mostly results in the diverse translation behaviors.  Only DLG$'$-BPE correctly translates it to the correct meaning ``\emph{how much}'', while FRQ$'$-BPE and AV$'$-BPE give the undesirable answers of ``\emph{how else}'' and ``\emph{what kind of}'' respectively. \end{CJK}

\section{Related Work}

Previous work resorted to various techniques to deal with the representation of rare words for NMT. Recently, character-level and other subword-based models have become increasingly popular and achieved great performance for different language pairs. The character-level models often rely on convolutional or recurrent neural networks to encode or decode the character sequences. For instance,  Costa-juss\`{a} and Fonollosa \cite{costajussa-fonollosa:2016:P16-2} employed convolutional and highway layers upon characters to form the word embeddings. Ling \etal \cite{LingTDB15} used a bidirectional LSTM to combine character embeddings to word embeddings and generated the target word character by character. Similarly,  Ataman and Federico \cite{ataman2018compositional} proposed to improve the quality of source representations of rare words by augmenting its embedding layer with a bi-RNN, which can learn compositional input representations at different levels of granularity. Luong and Manning \cite{luong-manning2016} proposed a word-character hybrid model that translates mostly at the word level and consults the character information for rare words with an additional deep character-based LSTM. Different from above methods which still consider word boundaries,  Chung \etal \cite{chung-cho-bengio2016} introduced a character-level decoder without explicit segmentation while the encoder is still at subword-level. Further, Lee \etal \cite{lee2017fully} proposed the fully character-level NMT model that maps a source character sequence to a target one without any explicit segmentation for both encoder and decoder. At present, the most popular subword-level method is independent of the training procedure and can be regarded as a preprocessing step such as BPE \cite{sennrich-haddow-birch} and the wordpieces \cite{wu2016google}. The main advantage of this approach is that the original architecture of the word-level model can be reused without increasing much complexity of training or decoding. Here, we focus on the subword segmentation improvement by introducing better substring measures.

\section{Conclusion}
In this paper, we introduce a generalized subword segmentation framework to enable open-vocabulary translation in NMT. Specifically, we extend the frequency only based BPE segmentation to a general bottom-up case that is capable of incorporating other advanced substring goodness measures, AV and DLG. We empirically compare the effects of different goodness measure schemes on the IWSLT14 German-English and WMT17 News Commentary Chinese-English translation tasks. The experimental results show that frequency weighted DLG$'$-BPE and AV$'$-BPE achieves stably better performance than FRQ$'$-BPE at different significant levels. In general, AV$'$-BPE shows best performance on the German-English task and DLG$'$-BPE
behaves better on the Chinese-English task. The difference on the best subword segmentation strategies indicates that the choice may be sensitive to specific language pairs to some extent, which deserves further exploration in the future.

\subsubsection*{Acknowledgments.}
This paper was partially supported by National Key Research and Development Program of China (No. 2017YFB0304100), National Natural Science Foundation of China (No. 61672343 and No. 61733011), Key Project of National Society Science Foundation of China (No. 15-ZDA041), The Art and Science Interdisciplinary Funds of Shanghai Jiao Tong University (No. 14JCRZ04).

\bibliography{ref}
\bibliographystyle{splncs04}

\end{document}